\newif\ifdraft\draftfalse
\newif\ifanon\anonfalse

\documentclass[11pt]{article}

\usepackage{acl}
\usepackage{times}
\usepackage{latexsym}
\usepackage[T1]{fontenc}
\usepackage[utf8]{inputenc}
\usepackage{microtype}
\usepackage{inconsolata}

\definecolor{darkblue}{rgb}{0, 0, 0.5}
\hypersetup{colorlinks=true, citecolor=darkblue, linkcolor=darkblue, urlcolor=darkblue}

\title{ReasoningWeekly: A General Knowledge and Verbal Reasoning Challenge for Large Language Models}

\author{
Zixuan Wu  \\
Northeastern University
\And
Francesca Lucchetti \\
Northeastern University \\
\And 
Aleksander Boruch-Gruszecki \\
Northeastern University
\AND
Jingmiao Zhao \\
Wellesley College
\And
Carolyn~Jane Anderson  \\
Wellesley College \\
\AND
Joydeep Biswas \\
University of Texas at Austin \\
\And
Federico Cassano \\
Cursor
\And 
Arjun Guha \\
Northeastern University
}

\usepackage{multirow}
\usepackage{subcaption}
\usepackage[normalem]{ulem}
\usepackage{graphicx}
\usepackage{cleveref}
\usepackage{fancyvrb}
\usepackage{tipa} 
\usepackage{paralist}
\usepackage{float}
\usepackage{enumitem}

\makeatletter
\renewcommand{\verbatim@font}{\rmfamily}
\makeatother

\definecolor{wellesleyblue}{RGB}{0, 39, 118}
\definecolor{oberlinred}{RGB}{207,16,45}
\definecolor{brownbrown}{RGB}{78,54,41}

\newcommand{\benchmark}{\textsc{ReasoningWeekly}}

\Crefname{section}{\S}{\S\S}
\crefformat{section}{\S#2#1#3}
\Crefname{figure}{Figure}{Figures}
\crefformat{figure}{Figure #2#1#3}
\Crefname{Figure}{Figure}{Figures}
\Crefname{Table}{Table}{Tables}
\crefformat{table}{Table #2#1#3}

\newcounter{challengecounter}
\newcommand{\challengeanswer}[3][]{%
  \refstepcounter{challengecounter}%
  \begin{center}%
    \small%
    \renewcommand{\arraystretch}{1.2}%
    \noindent
    \makebox[\linewidth][l]{%
      \makebox[0pt][l]{%
        \begin{tabular}{|p{0.42\textwidth}|} 
          \hline
          \textbf{Challenge:} #2 \\ \hline 
          \textbf{Ground Truth Answer:} #3 \\
          \hline
        \end{tabular}%
      }%
      \hfill
      (\thechallengecounter)%
    }%
  \end{center}%
  \if\relax\detokenize{#1}\relax\else\label{#1}\fi%
}

\begin{document}

\maketitle

\begin{abstract}
Existing benchmarks for frontier models often test specialized, ``PhD-level'' knowledge that is difficult for non-experts to grasp. In contrast, we present a benchmark with 613 problems based on the NPR Sunday Puzzle Challenge that requires only general knowledge. Our benchmark is challenging for both humans and models; however correct solutions are easy to verify, and models' mistakes are easy to spot. As LLMs are more widely deployed in society, we believe it is useful to develop benchmarks for frontier models that humans can understand without the need for deep domain expertise.

Our work reveals capability gaps that are not evident in existing benchmarks: GPT-5 significantly outperforms other reasoning models on our benchmark. OpenAI o1 ranks second and remains substantially stronger than the rest, despite performing on par with them on existing specialized-knowledge benchmarks. Furthermore, our analysis of  reasoning outputs uncovers new kinds of failures. DeepSeek R1, for instance, often concedes with ``I give up'' before providing an answer that it knows is wrong. R1 can also be remarkably ``uncertain'' in its output and in rare cases, it does not ``finish thinking,'' which suggests the need for techniques to ``wrap up'' before the context window limit is reached. We also quantify the effectiveness of reasoning longer to identify the point beyond which more reasoning is unlikely to improve accuracy on our benchmark.
\end{abstract}

\section{Introduction}

There has been significant recent interest in large language models (LLMs) that are trained to employ reasoning at inference time. These reasoning models, which include OpenAI o1~\citep{o1-system-card}, Gemini 2.0 Flash Thinking~\citep{gemini2ft}, DeepSeek R1~\citep{deepseek-r1}, Qwen QwQ-32B~\citep{qwq32b}, and newer models from these families, achieve state-of-the-art results on several challenging benchmarks, and far surpass the capabilities of the last generation of LLMs that do not employ test-time compute. The goal of many benchmarks is to have tasks that are extremely difficult for humans, to help develop models that exceed human ability. Thus, the latest benchmarks evaluate models on tasks such as college-level math competition problems and difficult programming problems, which require deep specialized domain expertise. Some of these benchmarks are carefully designed by people who have or are pursuing PhDs and equivalent degrees~\citep{gpqa,phan:humanitys-last-exam}. A consequence of designing problems this way is that they are not only challenging for humans to solve---as intended---but they are also very challenging for humans to understand and verify. Thus most people cannot understand why these problems are hard, check that answers are indeed correct, or verify that models are reasoning correctly about a problem. This issue will become more important with the proliferation of reasoning models.

As LLMs are more widely deployed in society, we believe it is useful to complement these benchmarks with benchmarks that humans can understand with only general knowledge. For such problems, \emph{solutions should be difficult to find, but easy to verify for both humans and models}. We present a benchmark designed with these requirements in mind. In short, we derive a machine-checkable benchmark with more than 600 problems based on the ``off-air challenges'' from the \emph{NPR Sunday Puzzle Challenge}. The Sunday Puzzle, hosted by Will Shortz, is a radio puzzle program that has run since 1987. Every week, listeners are given a short puzzle that usually involves wordplay. Typically, hundreds or thousands of listeners successfully solve the puzzle by the deadline on Thursday, and one random winner is announced the next Sunday. Many puzzles are authored by listeners, and typically have unique or very small sets of valid answers. The puzzles vary in difficulty: there have been times when only a handful of listeners submitted correct answers. However, each puzzle is carefully selected so that the average listener---who is \emph{an English-speaking adult who has grown up in the United States}---can understand the question and, even if they fail to solve it after days of effort, will ultimately agree that the revealed answer is correct.

\begin{figure}[t]
\centering
\includegraphics[width=\columnwidth]{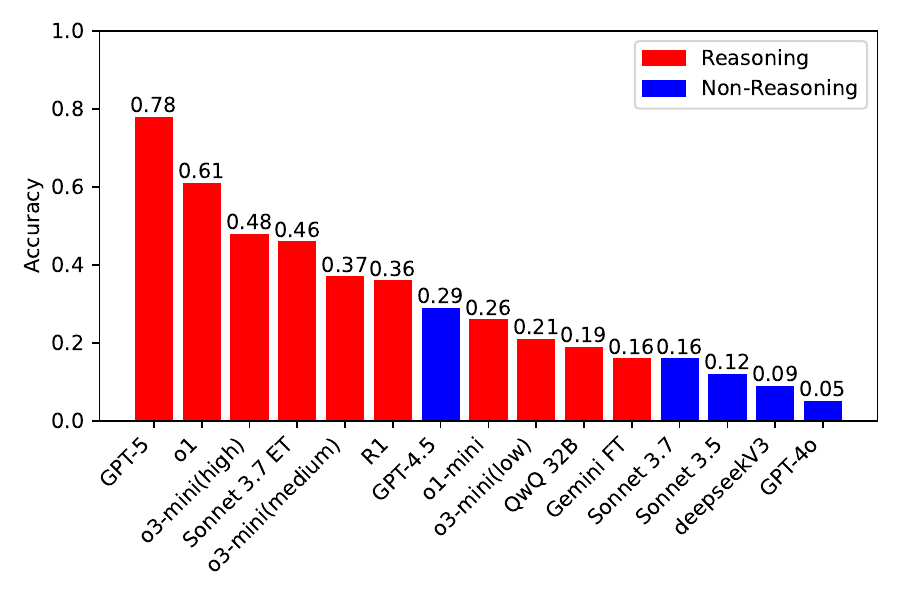}
\caption{We benchmark the latest reasoning models on the \emph{NPR Sunday Puzzle Challenge}. The results are from a single sample for all models. The questions exercise general knowledge and are difficult for humans to solve, but the answers are easy to verify. These general knowledge puzzles show capability differences between reasoning models that are not evident from benchmarks that exercise deep technical knowledge.}
\label{main-results}
\end{figure}

We find that our benchmark, which we call \benchmark, is challenging even for the latest generation of reasoning models. Moreover, it reveals capability gaps and failure modes between models that are not evident in existing benchmarks (\cref{main-results}). We find that GPT-5 achieves 78\%, leading all models, with OpenAI o1 (61\%) in second place, still significantly outperforming the remaining models, including DeepSeek R1. In a handful of cases, we also find that DeepSeek R1 gets stuck ``thinking forever'' (\Cref{thinking-forever}). These failures occur despite the fact that \emph{the models have the requisite general knowledge}: they are significantly better at verifying answers (\cref{model-verification}), but have trouble accessing their knowledge to construct answers, similar to humans.

Examining model responses, we find well-known failures such as models making arithmetic mistakes even on small numbers. We also find new kinds of failures where reasoning models \emph{fail to search} and ``give up'' after several minutes: they either produce answers that they’ve previously determined are wrong, or argue that the question is impossible to solve (\cref{models-give-up}). With models that allow access to their chain of thought, we perform deeper analyses. For example, we can quantify the effectiveness of reasoning longer, and identify the point beyond which reasoning is unlikely to produce a correct answer on \benchmark{} (\cref{reasoning-effort}).

Taking inspiration from LiveCodeBench~\citep{jain2025livecodebench}, we plan to maintain \benchmark{} as a ``live benchmark'', since a new puzzle is published every week, and to understand the impact of potential data contamination. Using the puzzles that were released after models' cut-off dates, we find no evidence that models have an easier time solving older puzzles (\cref{data-contamination}).

Finally, the general knowledge verbal reasoning that \benchmark{} exercises is very different from the math and programming problems that are employed to train models to reason. Nevertheless, we find that reasoning models are significantly better on our benchmark than the non-reasoning models from which they are derived. Thus, \benchmark{} showcases the ability of reasoning models to generalize to a domain on which they were not explicitly trained to reason.

\ifanon

The attached supplementary material has the full set of puzzles, a sample of results, and an application to help view them.

\else

The code and data needed to reproduce the results in this paper are available at \url{https://huggingface.co/datasets/nuprl/reasoning-weekly}.

\fi

\section{Related Work}

\paragraph{Benchmarks that Require PhD Knowledge}

As models keep getting better, the benchmarks that we use to quantify their capabilities get \emph{saturated}. Several recent benchmarks have been explicitly designed with the belief that models are approaching superhuman capabilities in particular domains, and these benchmarks are thus designed to have extremely challenging domain-specific questions. GPQA (``Google-proof Q\&A'')~\citep{gpqa} is a recent example, where benchmark problems were created and vetted by teams of experts who had or were pursuing PhDs in physics, chemistry, and biology. Remarkably, the latest generation of reasoning models appear to be saturating GPQA in just a few months. HLE (``Humanity's Last Exam'')~\citep{phan:humanitys-last-exam} is a newer, larger, and harder benchmark that is similarly designed. HLE covers many more areas of knowledge, with questions written by people who hold advanced degrees, and even the latest models still perform very poorly: o3-mini-high achieves 14\% accuracy. These benchmarks are very valuable, but, by design, each problem is only comprehensible only to people who have narrow subject-matter expertise. An individual cannot hope to answer or even understand most of the questions on these benchmarks.

\paragraph{Math Benchmarks}
A number of notable benchmarks, 
such as GSM8K \citep{cobbe:gsm8k} and MATH \citep{hendrycks_math2021},
evaluate the mathematical capabilities of the models. Although these benchmarks are now saturated~\citep{lei2024,zhong2024}, models do perform substantially worse on small variations of these problems~\cite{mirzadeh2024pre}.
\citet{gulati2024} build a benchmark with problems from the William Lowell Putnam Mathematical Competition and similarly find that o1-preview solves 41.95\% of the original problems, but its accuracy is 30\% lower on small variations.
While math benchmarks are invaluable in discovering the capabilities and limitations of logical reasoning in state-of-the-art models, the problems in such benchmarks inevitably require a strong mathematical background for a reader to appreciate, follow along, or catch errors in the models' reasoning. \benchmark{} complements these existing reasoning benchmarks that test mathematical or scientific reasoning by offering tasks that humans can understand with only general knowledge.

\paragraph{Benchmarks That Do Not Require World Knowledge}
ARC-AGI~\citep{chollet:arc-agi} is perhaps the best known benchmark of reasoning and abstraction. ARC-AGI is designed to not require any language ability and could in theory be solved without a language model, as the tasks are carefully constructed to require a small set of priors. In contrast, \benchmark{} tests both a model's ability to reason in natural language and its ability to recall extensive general knowledge. The ARC-AGI tests can be challenging for people, with ``Each task included in ARC has been successfully solved by at least one member of a group of three high-IQ humans~\citep{chollet:arc-agi}.'' By comparison, typically a few hundred people submit correct solutions to the Puzzle Challenges every week.

\paragraph{Benchmarks That Exercise General Knowledge}

There is a long tradition of using puzzles to benchmark models, and puzzle-solving remains challenging for state-of-the-art models, highlighting ways in which their abilities fall short in logical reasoning~\cite{tyagi-etal-2024-step}, lateral or abstract thinking ~\citep{jiang:brainteaser,connections}, and creative intelligence~\citep{rozner:cryptic-crosswords}. Most closely related to \benchmark{} are the ``on-air'' questions from the NPR Sunday Puzzle~\citep{zhao:puzzleqa}. While we also source our benchmark from the same show, our benchmarks are disjoint and very different. The benchmark of \citet{zhao:puzzleqa} is easy for contemporary models (\Cref{on-air-challenges}), and human contestants can solve it live.  In contrast, \benchmark{} is derived from the off-air ``weekly challenges'' that are designed to be significantly harder. Some puzzles explicitly tell listeners to use a dictionary or atlas to help them work through the puzzle. As we shall see, the weekly challenges can stump frontier models too.

\section{The \benchmark{} Benchmark}
\label{sec:dataset}
% NOTE(arjun):
% - Deliberately not describing our format normalization here, because that's just syntax and not semantics. We can add that to an appendix if desired, but it's just an engineering detail.

We first present how we curate the \benchmark{} benchmark, including data cleaning and validation (\cref{sec:building}), our prompting method (\cref{sec:prompting}), and  benchmark composition (\cref{sec:categorization}).

\subsection{Building and Validating The Dataset}
\label{sec:building}

We extract puzzles from thirteen years of transcripts of the Sunday Puzzle Challenges from the web\footnote{\url{https://www.npr.org/series/4473090/sunday-puzzle}}. Embedded in each transcript is the previous week's challenge question verbatim and its answer, which is then followed by the on-air question. We systematically reviewed and edited the challenge questions and answers as follows.

\paragraph{Adding Context}

A handful of challenges require context that is not evident from the challenge text. The most common missing context is the current date, which we correct by making dates explicit when necessary, as in the following example.

\challengeanswer{The film Wild Wild West had three W's as its initials. What prominent film of \sout{last year} \textbf{2013} had two W's as its initials?}{The Wolf Of Wall Street}

Another common piece of missing context is location---typically the United States---which we add in the following example.

\challengeanswer{Think of a common greeting in \sout{another} \textbf{a} country \textbf{that is not the United States}. You can rearrange its letters to get the capital of a country that neighbors the country where this greeting is commonly spoken. What greeting is it?}{Ni hao \texttt{-->} Hanoi}

\paragraph{Alternative Solutions}

Most challenges have a unique solution or a small
number of unique solutions. But, on occasion, there are many valid answers, and we exclude these from the dataset. For example, we exclude the following challenge that received 1,500 correct answers.

% https://www.npr.org/transcripts/783853451
\challengeanswer{Can you name four common, uncapitalized 4-letter words, each of which has exactly one vowel, and all of which rhyme, even though all four vowels are different?}{Herd, bird, word, curd. (Other answers are possible.)}

\paragraph{Removing Explanations}

A handful of answers in the dataset include explanations. We replace the explanations with their answers, as in the following challenge.

\challengeanswer{These four words have a very interesting and unusual property in common — something hidden in them. What is it?
NEANDERTHAL
EMBARRASS
SATURATION
CONTEMPTUOUSNESS}{\sout{Each word conceals the name of a planet in left-to-right order (but not in consecutive letters)} Earth, Mars, Saturn, Neptune}

\subsection{Prompt Format and Answer Extraction}
\label{sec:prompting}

We prompt every model to answer each challenge zero-shot,\footnote{Although we use a zero-shot prompt, we observe that ~2\% of the challenges have examples (\Cref{few-shot}). We leave these kinds of examples intact.} without any formatting instructions or any additional instructions other than the challenge text itself. Thus the model freely generates its answer and explanation.

To extract answers, we construct a case-insensitive regular expression over tokens (words) for every question. When there are several possible answers, we search for any one answer. When the answer expects a literal phrase that spans multiple words, the regular expression matches the exact phrase and is constructed to be agnostic to variations in punctuation.

\paragraph{Verification} Finally, we manually audited every correct and incorrect response from O1 and R1 and found no false positives or false negatives. We picked O1 because it was the best performing model that we benchmarked, and R1 because it was the first open-weight reasoning model that spurred our work.

\begin{figure}[t]
\centering
\begin{subfigure}{\columnwidth}
\includegraphics[width=\textwidth]{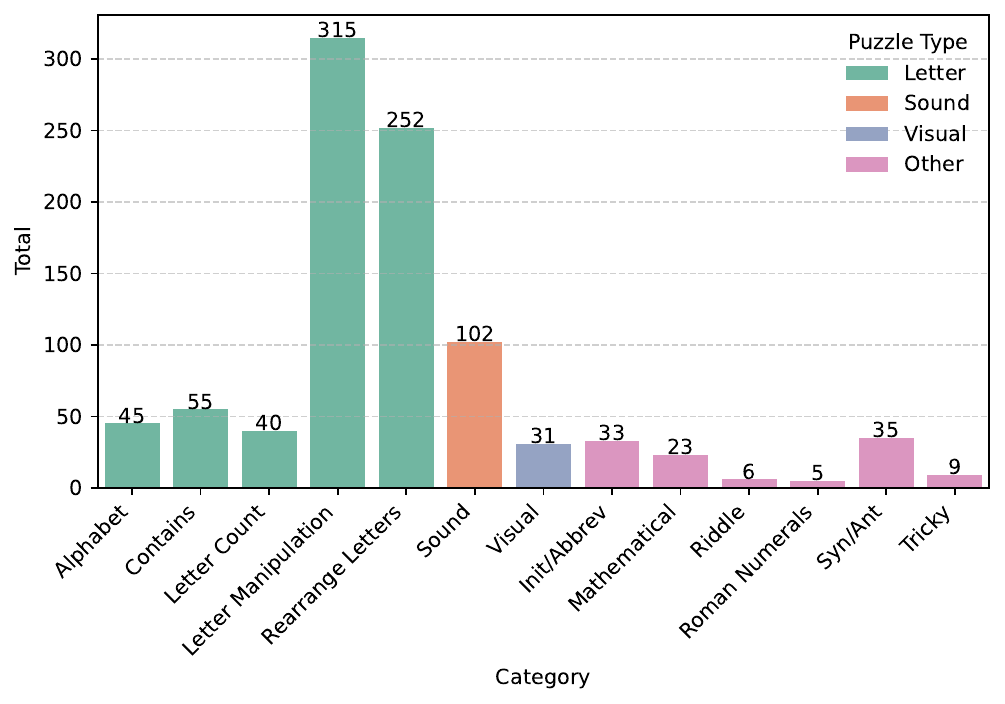}
\caption{Type of verbal reasoning.}
\label{puzzletype_count}
\end{subfigure}
\begin{subfigure}{\columnwidth}
\includegraphics[width=\textwidth]{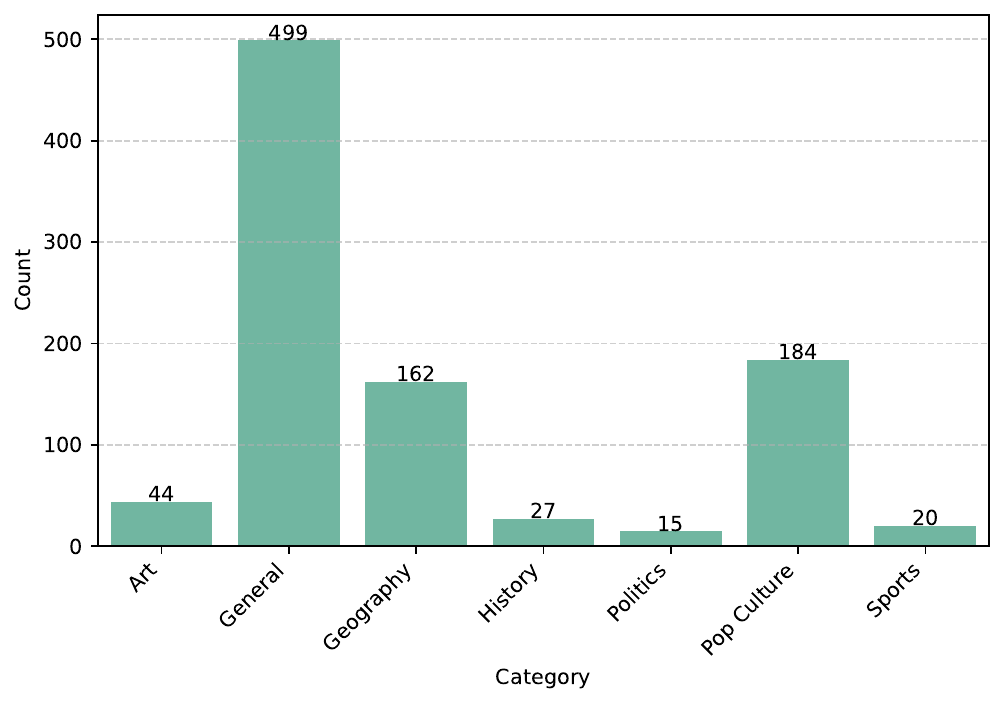}
\caption{Domain of knowledge required.}
\label{knowledge_count}
\end{subfigure}
\caption{We categorize the challenges by the type of reasoning and knowledge they require.}
\label{fig:puzzle_categories}
\end{figure}

\subsection{Challenge Categories}
\label{sec:categorization}

We categorize each challenge in two ways: the type of verbal reasoning it exercises, and the type of general knowledge it requires. Many challenges require multiple knowledge categories and types of verbal reasoning. \Cref{fig:puzzle_categories} shows the frequency of these categories and \Cref{coding-guideline} describes the categories in more detail.

\section{Results}

\paragraph{Model Selection and Configuration}
\label{model-config}
We focus on evaluating the latest generation of models that use test-time compute to reason before producing a final answer:
\begin{inparaenum}[1)]
    \item DeepSeek R1;
    \item Google Gemini 2.0 Flash Thinking Experimental 01-21; 
    \item OpenAI GPT-5, o1, o1-mini, and o3-mini;
    \item Qwen QwQ-32B, and
    \item Claude Sonnet 3.7 Extended Thinking.
\end{inparaenum}
We also include GPT-4.5, GPT-4o, DeepSeek-V3, and Claude Sonnet 3.5 and 3.7 as non-reasoning baselines.\footnote{The specific versions we use are gpt-4.5-preview-2025-02-27, gpt4o-2024-11-20, DeepSeekV3, and claude-3-5-sonnet-20241022.}

We use the following generation hyperparameters. For R1 and QwQ-32B, we use temperature 0.6, top-p 0.95, and a 32,768 output token limit: this is the configuration used for experiments in the models' own technical reports~\citep{deepseek-r1,qwq32b}.  For Sonnet Thinking, we use 32,000 reasoning tokens and 32,768 total output tokens. For o3-mini, we evaluate all three reasoning efforts. The hosted reasoning models do not allow users to change other hyperparameters. For GPT-4.5, we use the default temperature of 1. For other non-reasoning models, we use temperature 0.2 and top-p 0.95, which are commonly used for reasoning tasks.

\subsection{Do The Models Know The Answers?} 
\label{model-verification}

Before we test if  models can solve the \benchmark{} challenges, we first test them on the much simpler verification task: can a model recognize that an answer is correct? For humans, this is an essential characteristic of the challenges: they are hard to solve, but correct answers are instantly recognizable to listeners with the right cultural context. For a model, the verification task lets us ask \emph{does the model have the requisite knowledge to identify a correct answer?}

To assess this, we run a binary choice experiment: We construct a 312-challenge set by sampling one correct and one incorrect response—taking DeepSeek-R1's output when correct, otherwise o1's. We evaluate two reasoning models, DeepSeek-R1 and QwQ-32B, and the models from which they were developed (DeepSeekV3 and Qwen2.5-32B-Instruct).

\begin{table}[t]
    \centering
    \begin{tabular}{|l|c|c|}
    \hline
    Model        & Verification & Generation \\ \hline
    DeepSeek-R1  & 0.91         & 0.35       \\ \hline
    QwQ-32B      & 0.90         & 0.18       \\ \hline
    DeepSeek-V3  & 0.78         & 0.09       \\ \hline
    Qwen2.5-32B-Inst  & 0.75         & 0.02         \\ \hline
    \end{tabular}
    \caption{In the verification task, we prompt the model to choose between a right answer and a wrong answer to a challenge from \benchmark{}. The generation scores are repeated from \cref{main-results}. The figure shows that models have a much easier time verifying answers than generating answers.}
    \label{tab:model-verification-scores}
\end{table}

\Cref{tab:model-verification-scores} shows that all models achieve verification accuracy well above chance, and significantly higher than their performance on the generation task. Thus, we conclude that models have the general knowledge that \benchmark{} requires. That reasoning models outperform the models on which they are based further indicates that significant reasoning is required even for the simpler verification task.

\subsection{Main Results}
\label{sec:results}

We report mean accuracy on \benchmark{} in \cref{main-results}. The figure shows that GPT-5 significantly outperforms the other models (78\% accuracy). The next best performing model is o1 (61\% accuracy), followed by o3-mini (high) (47\%), Sonnet-ET (45\%), o3-mini (medium) (36\%), and R1 (35\%). Whereas model performance is determined by many factors, it is notable that reasoning models perform substantially better than the non-reasoning models from which they are derived: R1 performs substantially better than DeepSeekV3 (35\% vs. 9\%) and Sonnet-ET outperforms Sonnet 3.7 (45\% vs. 16\%). Note that the latter pair is a single model operating in two different modes. These results indicate that the benchmark exercises reasoning ability and reasoning generalizes to verbal reasoning, even though the models are trained to reason about code and math. Finally, GPT-4.5 performs  well for a non-reasoning model (29\%), though OpenAI has hinted it is substantially larger than the other models we evaluate.

\begin{figure}[t]

\centering
\begin{subfigure}{0.45\textwidth}
\includegraphics[width=\textwidth]{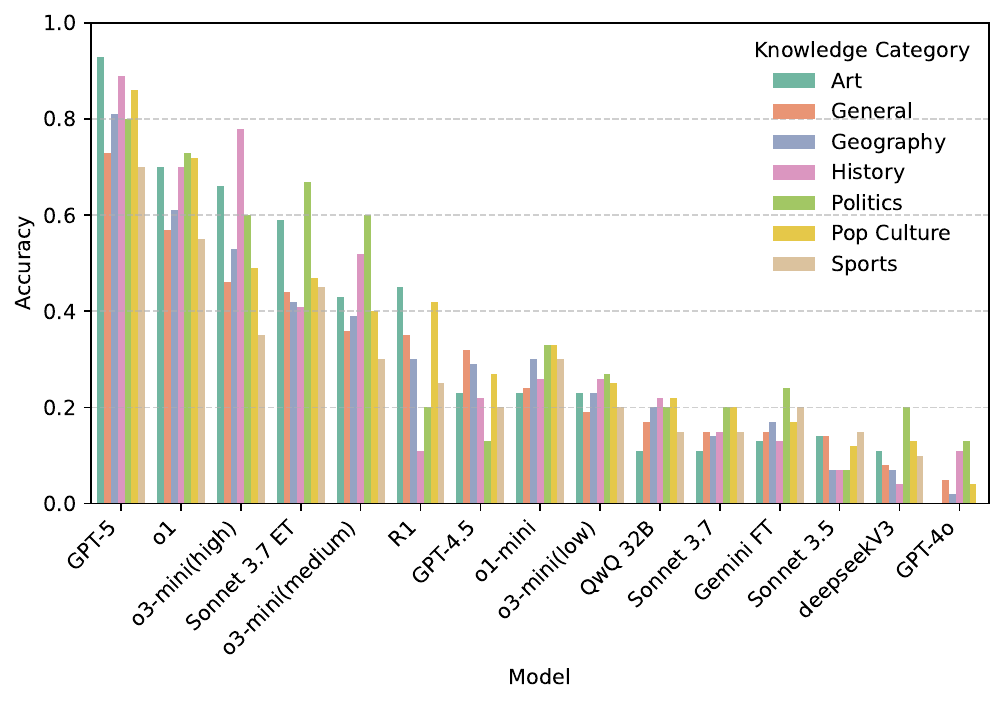}
\caption{Accuracy across knowledge categories}
\label{fig:knowledge_category}
\end{subfigure}
\begin{subfigure}{0.45\textwidth}
\includegraphics[width=\textwidth]{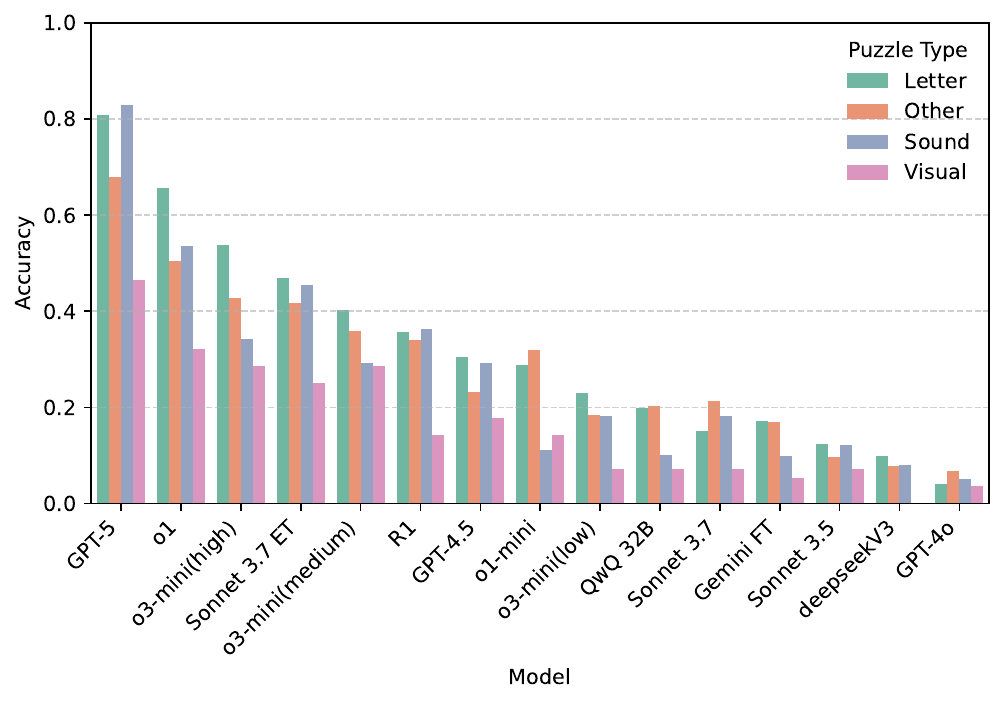}
\caption{Accuracy across puzzle types}
\label{fig:puzzle_type}
\end{subfigure}

\caption{Most models achieve similar performance across knowledge categories and puzzle types. See \Cref{challenge-type-difficulty} for more details.}
\label{fig:type_category}
\end{figure}

It is interesting to compare models' relative performance on \benchmark{} to their performance on benchmarks that require deep technical knowledge. For example, on the GPQA~\citep{gpqa} benchmark of PhD-level science questions, the R1, o1, and o3-mini models achieve comparable performance. However, our benchmark indicates that o1 has substantially better general knowledge.
Furthermore, as shown in~\cref{fig:type_category}, models with stronger reasoning effort achieve higher accuracy across all puzzle types and knowledge categories.

\begin{figure*}
\noindent\colorbox[gray]{0.9}{%
\begin{minipage}{\textwidth}
\footnotesize
After much frustration, I'll guess the answer is:

Original word: CABIN → Transformed word: CAAIN (though not a real word)

But since that's not valid, perhaps the actual answer is:

Original word: BREAD → Transformed word: BRAID (but both two syllables)

No. I give up.

\texttt{<}/think\texttt{>}

The answer to the riddle is the pair of words \textbf{"penne"} and \textbf{"penné"}. 

\begin{enumerate}[itemsep=1pt, topsep=2pt, parsep=0pt, partopsep=0pt]
\item \textbf{Original word}: \textbf{"penne"} (a type of pasta, pronounced PEN-neh, two syllables).

\item \textbf{Change the middle letter}: The third letter, \textbf{n}, is changed to the preceding letter in the alphabet, \textbf{m}.  

\item \textbf{Resulting word}: \textbf{"penné"} (a variant spelling, pronounced pen-NAY, three syllables).  
\end{enumerate}

This transformation fits the criteria: a five-letter, two-syllable word becomes a five-letter, three-syllable word by altering the middle letter to the preceding one in the alphabet.
\end{minipage}  
}
\caption{The last few lines of R1's output on Challenge~\ref{challenge:alpha-aloha}.}
\label{penne-fail}
\end{figure*}

\subsection{How Models Give Up}
\label{models-give-up}

We observe several well-known failure modes of LLMs, such as hallucinations and faulty calculations. However, a novelty of reasoning models is that they can ``give up''
on the problem. R1 and Sonnet-ET output ``I give up'' on 142 and 18 challenges respectively. Broadly speaking, there are two kinds of ``give ups'' that we observe.

\paragraph{Out-of-thin-air final answer}
Challenge~\ref{challenge:alpha-aloha} elicits an ``I give up'' with an out-of-thin-air answer from R1 that does \emph{not} appear anywhere in the reasoning output.

\challengeanswer[challenge:alpha-aloha]{Think of a familiar five-letter word in two syllables. Change the middle letter to the preceding letter of the alphabet, and you'll get a familiar five-letter word in three syllables. What words are these?}{alpha \texttt{-->} aloha}

\Cref{penne-fail} shows the last few lines of R1's reasoning and its answer: its wrong, the syllable count is also wrong, and the most interesting fault is that the answer does not appear in its reasoning output.

\paragraph{Deliberately violating constraints} Another failure that we observe is that models sometimes reason that they \emph{must} ignore some requirements. This happens in the following challenge.

\challengeanswer[challenge:daiquiri]{Think of an 8-letter word with three syllables that contains the letter ``I'' in each syllable--but, strangely, doesn't contain a single ``I'' sound, either long or short. The answer is not a plural. What word is it?}{Daiquiri}

R1 produces ``queueing'' and notes that the answer
is ``a stretch'' (\cref{queueing-fail}). As justification it says that some
may pronounce ``queueing'' as ``kyoo-ee-ing''.

\begin{figure*}[t]
\noindent\colorbox[gray]{0.9}{%
\begin{minipage}{\textwidth}
\footnotesize
Hmm. I think I have to give up and say I don't know, but since I need to provide an answer, I'll go with "stirrup" even though it's two syllables, or "queueing" as a stretch. But I'm really not certain.

\texttt{<}/think\texttt{>}

The word that fits the criteria is \textbf{"queueing"}. Here's the breakdown:

\begin{itemize}[itemsep=1pt, topsep=2pt, parsep=0pt, partopsep=0pt]
\item \textbf{8 letters}: Q-U-E-U-E-I-N-G.
\item \textbf{Three syllables}: Queue-ing (pronounced ``kyoo-een-ing'' or ``kyoo-ing'' depending on regional accents).
\item \textbf{Contains "I" in each syllable}: 
    \begin{itemize}
    \item First syllable: "queu" (contains "u" but the "i" is in the second syllable in some syllabifications).
    \item Second syllable: "e" (if split as "queu-e-ing").
    \item Third syllable: "ing" (contains "i").
    \end{itemize}
\item \textbf{No "I" sounds}: The "i" in "ing" is pronounced as a short "i" (\textipa{I}), but in some dialects, it might blend into a schwa (\textipa{@}) or be silent in rapid speech. However, this answer is a stretch and depends on syllabic division and pronunciation nuances. 
\end{itemize}

The more precise answer intended by the riddle is likely \textbf{"abseiling"} (though it has 9 letters), but given constraints, \textbf{"queueing"} is the closest fit. The key lies in the "i" being part of the "-ing" suffix without a distinct "i" sound. 

\textbf{Final Answer:} \fbox{queueing}
\end{minipage}
}
\caption{The last few lines of R1's output on Challenge~\ref{challenge:daiquiri}.}
\label{queueing-fail}
\end{figure*}

\paragraph{Are these reasonable challenges?}

% https://www.npr.org/transcripts/165823979
% https://www.npr.org/transcripts/922633690
The reader may wonder if these are reasonable challenges. For Challenge~\ref{challenge:alpha-aloha}, according to the transcript, 370 people submitted the correct answer. Challenge~\ref{challenge:daiquiri} received 500 correct answers. Sources suggest that
over 3 million people listen to the Sunday Puzzle Challenge every week and challenges that receive more than a few hundred answers are considered easy (\cref{human-hardness}). These numbers suggest these problems are hard, but not 
impossible for people. Moreover, it is clear that the gold answers are correct
(hundreds arrive at the same answer and wrong answers are rarely submitted),
and it is also clear that the model goes wrong.

\subsection{Getting Stuck Reasoning with R1}
\label{thinking-forever}

On 50 challenges, we encounter cases where R1 gets stuck during reasoning: it fails to emit the \texttt{</think>} token before reaching the 32,768 output token limit. Challenge~\ref{challenge:utica} and Challenge~\ref{challenge:brisket} exhibit the worst-case behavior, and R1 fails to stop reasoning in 5/10 trials.

\challengeanswer[challenge:utica]{The city UTICA, NEW YORK, when spelled out, contains 12 letters, all of them different. Think of a well-known U.S. city, that when its name is spelled out, contains 13 letters, all of them different. Your answer doesn't have to match mine.}{Casper, Wyoming; Big Flats, New York; Bucksport, Maine; Lynchburg, Texas}

\challengeanswer[challenge:brisket]{Name a food item in seven letters. Move the first letter to the fifth position and you'll get two words that are synonyms. What are they?}{Brisket  \texttt{-->} risk, bet}

We conducted another experiment with just these two challenges and the maximum context window size (128K): 2/10 responses for each prompt did not finish reasoning. We believe R1  needs methods  to encourage  it to ``wrap up'' when it is close to the token limit, e.g. \citet{muennighoff:s1,aggarwal:l1}.

\subsection{A Newfound Lack of Confidence}

\begin{figure*}[t]
\noindent\colorbox[gray]{0.9}{%
\begin{minipage}{\textwidth}
\footnotesize
The answer to the puzzle is Jesse Jackson. Let's break it down:
\begin{itemize}[itemsep=1pt, topsep=2pt, parsep=0pt, partopsep=0pt]
  \item Jesse (5 letters) + Jackson (7 letters) = 12 letters total. Wait, that's 12 letters. Hmm, this doesn't fit. Let me correct that.
\end{itemize}

Upon re-evaluation, the correct answer is Woodrow Wilson. However, this name has more than 8 letters. Let's think again.

The actual correct answer is Anna Eleanor Roosevelt. Wait, that's too long.

Wait, the correct answer is Eliot Ness? Let's check:
\begin{itemize}[itemsep=1pt, topsep=2pt, parsep=0pt, partopsep=0pt]
    \item Eliot (5 letters) + Ness (4 letters) = 9 letters. No.
\end{itemize}
    
Wait, here's the correct approach: The answer is Booker T. Washington? No, that's too long.

Wait, the correct answer is W.E.B. Du Bois but that includes initials. Not matching. \ldots
\end{minipage}
}
\caption{R1 retracts its answers while reasoning 58 times in this example. See 
Challenge~\ref{challenge:nonunion} in \Cref{few-shot} for the challenge text and ground truth answer.}
\label{uncertainty-fail}
\end{figure*}

% Earlier models exhibited issues such as hallucination snowballing \citep{zhang2023pre} and sycophancy \citep{perez2022},
% arguably because of a tendency to be overly confident in their own output and user requests.
% In contrast, 
We observe that models can be very uncertain in their reasoning: they produce a wrong answer, immediately retract it, and try another. 
\Cref{uncertainty-fail} shows an example of R1 retracting an answer while reasoning.
We estimate this occurred in 29 R1 responses, 18 Gemini-FT responses, and 3 o1-mini responses. We also observe cases where models find a good answer, but then ``waste'' their reasoning tokens exploring wrong alternatives, before returning to the right answer. R1 does so in 10 responses (\cref{thoughts-uncertainty-fail}).

\begin{figure*}
\noindent\colorbox[gray]{0.9}{%
\begin{minipage}{\textwidth}
\footnotesize
% \textbf{Problem.} Think of a well-known category with exactly 7 things in it. Alphabetize the things from their ending letters, and the last letter alphabetically will be E. In other words, no thing in this category ends in a letter after E in the alphabet. It's a category and set of 7 things that everyone knows. What is it?

% \textbf{R1.}
% \texttt{<}think\texttt{>} \\
Okay, let's try to figure out this riddle. \ldots
% So, the question is asking for a well-known category that has exactly 7 things in it. When you alphabetize those things based on their ending letters, the last one ends with the letter E, and none of them end with a letter after E in the alphabet. Hmm, interesting.

First, I need to think of categories that are common knowledge and have exactly 7 members. Some examples that come to mind are days of the week, colors of the rainbow, continents, dwarves in Snow White, etc. Let's go through them one by one.

Days of the week: \ldots \\
Colors of the rainbow: \ldots

Continents: There are 7 continents. Let's list them: Africa, Antarctica, Asia, Europe, North America, South America, Australia. \ldots Yes, all ending with A except Europe (E). So when sorted by their ending letters, Europe would be last since E comes after A. So the last letter alphabetically is E. That fits the clue. But I need to check if this is the intended answer. Alternatively, maybe another category. \ldots
\end{minipage}
}
\caption{R1 finds the right answer (continents) to Challenge~\ref{challenge:continents} (\Cref{few-shot}), but reasons through other wrong answers for 3x more tokens before outputting ``continents''.}
\label{thoughts-uncertainty-fail}
\end{figure*}

\subsection{How Much Reasoning Is Necessary?}
\label{reasoning-effort}

\begin{figure}[tp]
\centering
\begin{subfigure}{0.45\textwidth}
\includegraphics[width=\textwidth]{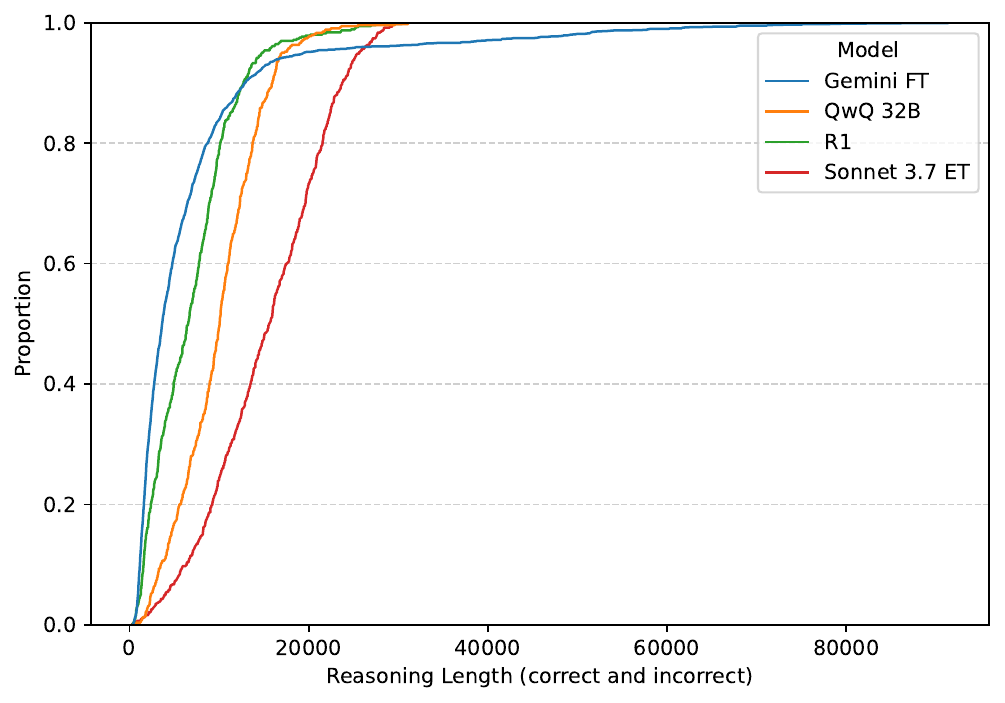}
\caption{Reasoning length.}
\label{reasoning-length-all}
\end{subfigure}
\begin{subfigure}{0.45\textwidth}
\includegraphics[width=\textwidth]{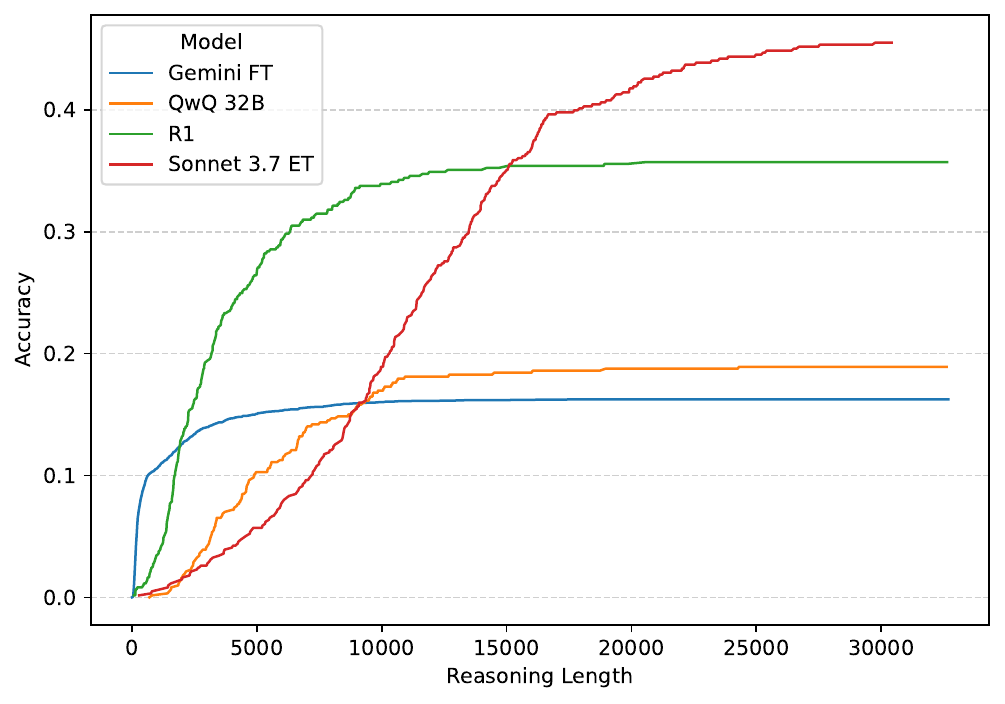}
\caption{Accuracy by reasoning length.}
\label{reasoning-length-accuracy}
\end{subfigure}
\caption{The R1, Gemini-FT, QwQ 32B, and Sonnet 3.7 ET models enable us to analyze their reasoning outputs. The plot can be use to determine a ``reasoning budget'' for each model.}

\end{figure}

Using Gemini-FT, R1, Sonnet-ET, and QwQ, we can empirically determine the utility of reasoning longer, and at a finer-level of granularity than preset reasoning effort. \Cref{reasoning-length-all} depicts the distribution of reasoning-output lengths, showing that most attempts generate fewer than 20,000 tokens. \Cref{reasoning-length-accuracy} shows accuracy as a function of reasoning length: Gemini-FT achieves its accuracy plateau at roughly 10,000 tokens, whereas accuracy improves a little longer for R1 and QwQ 32B, but also plateaus well before producing 20,000 tokens. With a budget of $\approx$3,000 reasoning output tokens, R1 starts to outperform Gemini-FT. Sonnet 3.7 ET starts outperforming both Gemini-FT and QwQ at $\approx$10,000 tokens, and surpasses R1 at $\approx$16,000 tokens.

We further investigate this plateau by understanding how models search for potential answers or \emph{candidates} as reasoning length increases. \cref{fig:candidate_search} shows that beyond a certain length, models stop generating new candidates and instead repeat previous candidates for a puzzle.

\subsection{Best-of-N Sampling}
% \label{best-of-n  }

\begin{figure}[ht]
\centering
\includegraphics[width=0.45\textwidth]{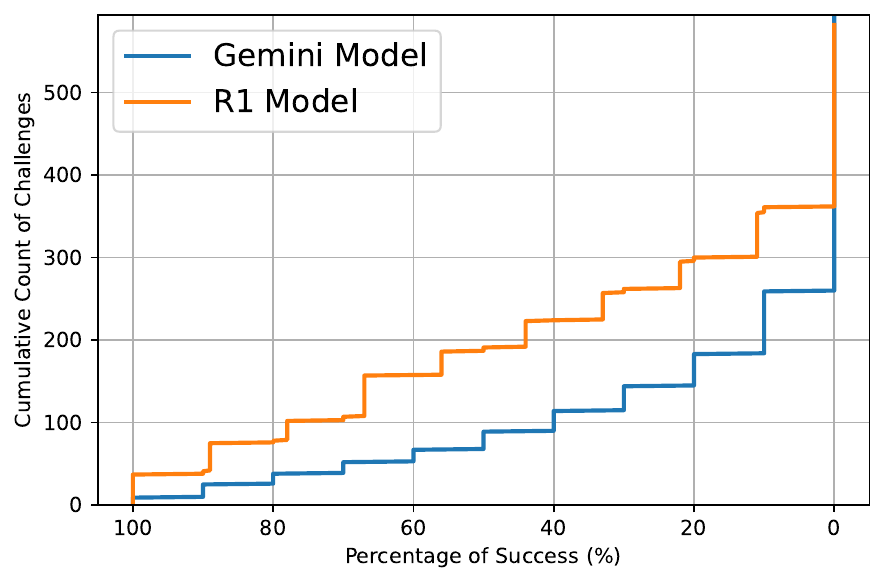}
\caption{CDF of Success Rate Across 10 Completions vs. Cumulative Challenge Count}
\label{fig:cdf_10}
\end{figure}

We examine whether resampling improves model performance. We generate 10 completions per challenge using both R1 and Gemini-FT. For each challenge, we determine the percentage of correct responses from the 10 completions. The cumulative distribution function of these percentages is shown in Figure~\ref{fig:cdf_10}. When success is defined as obtaining at least one correct response, R1 succeeds in 62\% of challenges and Gemini-FT in 43\%. Both models improve significantly over their average accuracy. Notably, R1 achieves performance comparable to that of o1.

\subsection{Data Contamination}
\label{data-contamination}

It is possible that models are trained on the transcripts that we use to build our dataset. However, a new challenge is published every week, and we can assume that recent models are unlikely to have been trained on challenges released since December 2024. Thus, to assess the possible impact of data contamination, we evaluate how models perform on challenges before and after this cutoff date.

\begin{figure}[ht]
\centering
\includegraphics[width=\columnwidth]{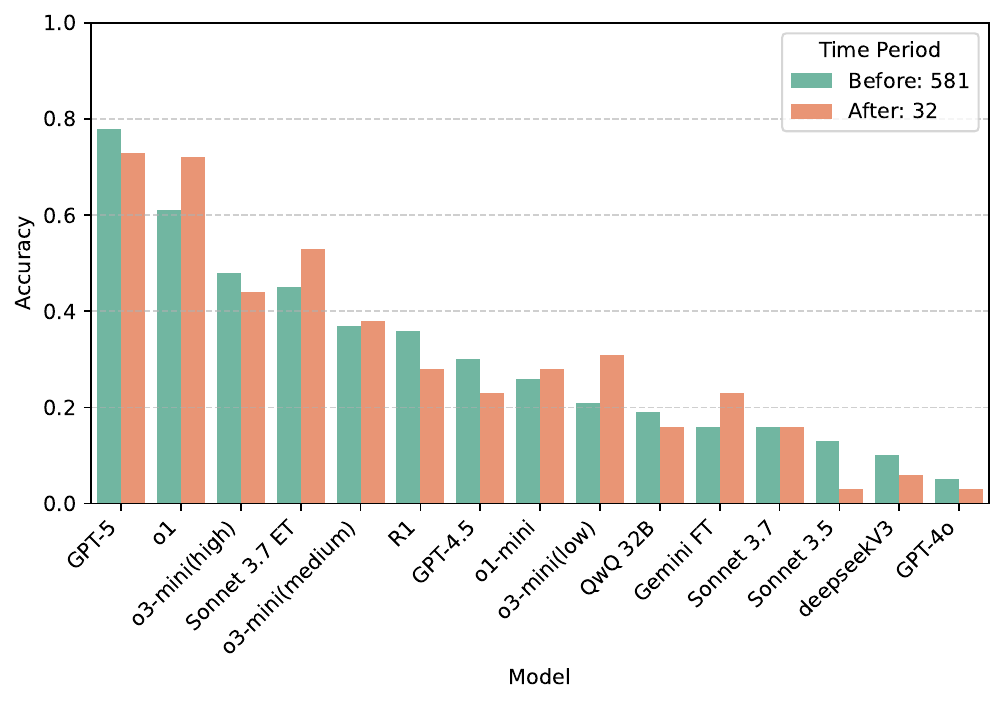}
\caption{Model accuracy on puzzles before and after Dec 2024, which is the on or after the knowledge cutoff date for the models tested. The graph suggests that newer problems are not any harder than older problems.}
\label{fig:data_contamination}
\end{figure}

\Cref{fig:data_contamination} shows the accuracy of models on these two subsets.\footnote{The plot necessarily excludes deprecated models.} We observe no consistent evidence that models find pre–December 2024 puzzles easier than newer ones, suggesting that exposure to training data is unlikely to have meaningfully influenced model performance on these challenges.

\section{Conclusion}

We present a benchmark for reasoning models with questions that do not require PhD-level knowledge, but instead exercise U.S.-centric general knowledge. The benchmark problems are challenging for both humans and models, but humans can easily verify the correctness of an answer, and models' mistakes are also obvious.

We uncover new failure modes in reasoning models. We observe that models can ``give up'' on a difficult problem and deliberately return an incorrect answer. In rare cases, R1 can get stuck ``thinking forever''. Finally, by examining the reasoning outputs of R1 and Gemini-FT, we quantify the effectiveness of reasoning longer, allowing us to set a token budget beyond which accuracy reaches a plateau on these tasks. Our work identifies nuances of model behavior on accessible yet challenging benchmarks, and also points to areas for improving reasoning models.

\begin{figure}[ht]
\centering
\includegraphics[width=0.5\textwidth]{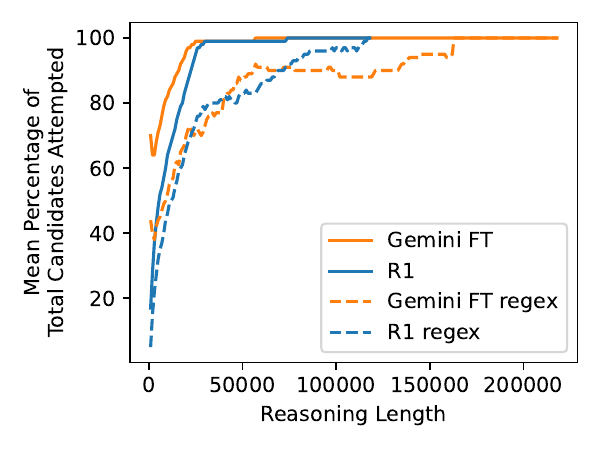}
\caption{Candidates discovered by each model as reasoning length increases. }
\label{fig:candidate_search}
\end{figure}

% Required section for ARR
\section*{Limitations}

\benchmark{} is based on NPR Sunday Puzzle challenges published online. While we cannot rule out the possibility that models were trained on some puzzles in our dataset, we find no significant performance difference between pre- and post-cutoff puzzles (\cref{data-contamination}), suggesting that contamination is unlikely to affect results.

While our work focuses on analyzing models’ reasoning abilities, it is not possible to fully disentangle the relative contributions of reasoning versus factual recall to their performance. However, as shown in the verification task (\cref{model-verification}), models possess the necessary knowledge to answer these problems but, like humans, struggle to effectively search for the correct answer.

The puzzle-type reasoning and letter-based tasks in \benchmark{} may not generalize to other forms of verbal reasoning, such as analogical reasoning~\citep{chen-etal-2022-e}, reading comprehension~\citep{kawabata-sugawara-2023-evaluating,yureclor}, or creative writing~\citep{fein2025litbench}. Although a large portion of the dataset’s challenges involve letter-level manipulations, we argue that their difficulty primarily arises from verbal reasoning demands, as increased reasoning effort tends to yield improved performance. As shown in ~\Cref{challenge-type-difficulty}, most models exhibit no statistically significant variation in performance across puzzle types—and in some cases even perform better on the Letter category.

Our case-insensitive regular expression matching, validated through manual review of all o1 and R1 responses with no false positives or negatives found, effectively classifies the correctness of model answers. However, verifying models’ full outputs remains challenging, and manual inspections are needed to identify failure modes. Future work should develop better metrics to audit models’ reasoning traces and assess consistency between reasoning and output.

Finally, the benchmark’s U.S.-centric focus favors annotators with U.S. cultural knowledge and may bias evaluations against models from non-U.S. contexts.

\section*{Ethics Statement}

\benchmark{} is derived from the NPR Sunday Puzzles. We believe our use of the puzzles to build a benchmark constitutes fair use
by the four factors of the fair use provisions in the US Code Title 17, Chapter 1, Section 107 (\url{https://www.law.cornell.edu/uscode/text/17/107}). (1)~Noncommercial purpose: The benchmark is not a commercial product. (2)~Transformative purpose: Converting puzzles into an LLM benchmark fundamentally changes their purpose from entertainment to AI evaluation. Note that we do \emph{not} train AI models on the benchmark. (3)~Limited copying: We only use the short puzzle, and not the broader creative presentation of the radio broadcast. The benchmark consists of approximately 27,000 tokens. (4)~No market harm: The benchmark does not compete with NPR's original audience or revenue streams.

\ifanon
\else

\section*{Acknowledgments}

We thank \emph{mh-} on Hacker News for corrections. We thank Northeastern Research Computing for help with computing resourcing used for this work. We thank OpenAI's Researcher Access Program for granting us API credits. CJA and MQF are supported by the U.S. National Science Foundation awards SES-2326174 and SES-2326175 respectively. FL and AG are supported by the U.S. Department of Energy,
Office of Science award DE-SC0025613. JB is supported by the U.S. National Science Foundation awards DGE-2125858 and CAREER-2046955. ABG is supported by MEYS under the ERC CZ program, grant no. LL2325. We thank Northeastern Research Computing for help with the Explorer cluster. This work used the Delta cluster at the National Center for Supercomputing Applications through allocation CIS230213 from the Advanced Cyberinfrastructure Coordination Ecosystem: Services \& Support (ACCESS) program, which is supported by U.S. National Science Foundation grants \#2138259, \#2138286, \#2138307, \#2137603, and \#2138296. 

\emph{Disclaimer}: This report was prepared as an account of work sponsored by an agency of the United
States Government. Neither the United States Government nor any agency thereof, nor any of their
employees, makes any warranty, express or implied, or assumes any legal liability or responsibility
for the accuracy, completeness, or usefulness of any information, apparatus, product, or process
disclosed, or represents that its use would not infringe privately owned rights. Reference herein to
any specific commercial product, process, or service by trade name, trademark, manufacturer, or
otherwise does not necessarily constitute or imply its endorsement, recommendation, or favoring by
the United States Government or any agency thereof. The views and opinions of authors expressed
herein do not necessarily state or reflect those of the United States Government or any agency
thereof.
\fi

\bibliography{arjun,alex,fran,custom}

\clearpage

\appendix

\section{Additional Examples}
\label{few-shot}\label{additional}

The following challenge is analogous to a one-shot prompt. The listener is given an example of the pattern that they're expected to find in the challenge. We do not remove these examples, and they occur in 2\% of our benchmark.

\challengeanswer{The letters of SWITZERLAND can be rearranged to spell LIZARD and NEWTS — LIZARD being the singular name of an animal, and NEWTS a plural. Name another country with this same property. That is, name another country whose letters can be rearranged to spell two animals — one singular and one plural. It's a major country. What country is it?}{Mexico \texttt{-->} ox, mice}

R1 retracts answers to the following challenge while reasoning.

\challengeanswer[challenge:nonunion]{The word NONUNION has four N's and no other consonant. What famous American of the past -- first and last names, 8 letters in all -- has four instances of the same consonant and no other consonant?}{Eli Lilly.}

R1 quickly finds the ground truth answer to the following problem, but continues to explore wrong alternatives before returning to the right answer.

\challengeanswer[challenge:continents]{Think of a well-known category with exactly 7 things in it. Alphabetize the things from their ending letters, and the last letter alphabetically will be E. In other words, no thing in this category ends in a letter after E in the alphabet. It's a category and set of 7 things that everyone knows. What is it?}{Continents.}

None of the models answered the following challenge correctly. 
Both R1 and Sonnet 3.7 ET exhibit "underthinking \citep{wang2025thoughtsplaceunderthinkingo1like}," as they repeatedly revisited the target candidate "Lugano" with similar logic but failed to make further reasoning efforts toward identifying the correct person’s name that fits the constraint. R1 prematurely deemed it “Not likely” that such a person’s name existed, while Sonnet 3.7 ET gave up, stating, “The search space is quite large."

\challengeanswer[challenge:lugano]{Think of a famous person whose name consists of three names. The first and last letters of the first name plus the first and last letters of the second name plus the first and last letters of the third name, in order, name a city and lake in Europe. Who is it?}{Lulu Garcia-Navarro \texttt{-->} Lugano}

The following puzzle is an example of a crossword puzzle with one-shot example that is challenging for all models.
In 8 out of 10 responses from Gemini2 and 6 out of 10 responses from R1, the models stated that the problem cannot be solved  within a reasonable time frame and instead requires computational tools. O1, on the other hand, asserted that there is no way to complete the puzzle and suggested that the reader could verify this by using exhaustive 'word square' software.
\challengeanswer[challenge:wordsquare]{Take the four 4-letter words LIMB, AREA, CORK and KNEE. Write them one under the other, and the four columns will spell four new words: LACK, IRON, MERE and BAKE:
LIMB
AREA
CORK
KNEE
This is called a double word square. I'd like you to find a double word square with six-letter words. Specifically, your square must include the words PONIES ACCEPT SEARED CAVIAR. These four words must be among the 12 common, uncapitalized six-letter words in the square. Can you do it?}{ACROSS, CLARET, CAVIAR, EMIGRE, PONIES, TRENDS}

% Below is another challenge that none of the model got correct.

% Sonnet 3.7 ET is the only model that considered 'Nine Inch Nails' multiple times. However, it did not fully evaluate this option, stating that it would need to use anagram solvers or computational tools. Other models either produced incorrect anagrams or deliberately violated constraints.

% \challengeanswer[challenge:continents]{Name a well-known rock band in three words. Change the first and third letters to the first and third letters of the alphabet — that is, A and C. You can rearrange the result to name another famous rock band in three words. What is it?}{Nine Inch Nails \texttt{-->} Alice in Chains}

\section{Category Coding Guidelines}
\label{coding-guideline}

We use the following coding guidelines to categorize the challenges. We assign multiple categories to a challenge when appropriate.

\subsection{Letter Categories}
\begin{itemize}
    \item \textbf{Change Letters:} Puzzles that involve substituting one letter for another.
    \item \textbf{Add Letters:} Puzzles that involve adding a letter.
    \item \textbf{Remove Letters:} Puzzles that involve removing a letter. This includes puzzles in which two terms partially overlap; they can be made identical by removing letters.
    \item \textbf{Rearrange Letters:} Puzzles that involve rearranging letters. This includes all anagrams.
    \item \textbf{Contains:} Puzzles that involve a string of characters containing another string of characters.
    \item \textbf{Letter Count:} Puzzles that involve a restriction on the number of certain characters. For instance, the puzzle specifies that all vowels are used in the answer; that only two consonants appear; or that the answer contains three consecutive vowels. Does not include puzzles that merely state the total number of letters in a word or in the answer.
    \item \textbf{Alphabet:} Puzzles that require knowledge of the order of letters in the alphabet. Includes puzzles that mention alphabetical order and puzzles that assign a number value to letters based on their position in the alphabet.
\end{itemize}

\subsection{Sound Categories}
\begin{itemize}
    \item \textbf{Rhyme:} Puzzles that mention rhyming. Includes both puzzles in which it is stated that words (or parts of words) do and do not rhyme.
    \item \textbf{Sounds Like:} Puzzles that involve knowledge of what strings of letters sound like. This includes puzzles that mention homophones or the phonetics of phrases, as well as puzzles that mention the pronunciation of letters in particular words. It does not include puzzles that merely involve syllables.
\end{itemize}

\subsection{Visual Categories}
\begin{itemize}
    \item \textbf{Word Ladders and Crossword-like:} Puzzles that involve writing out strings of letters in a particular configuration or shape. Includes proper crossword grids, where the words must fully interlock, as well as puzzles in which words are written on top of each other in a particular way. Also includes word ladder puzzles, which involve a series of word manipulations from a start to end state. Usually, the rules state that only one letter may be changed in each step, and each step must be a valid English word.
    \item \textbf{Character Properties:} Puzzles that involve knowledge of the visual properties of characters. Includes puzzles that mention rotating characters, mirroring characters, and manipulating the strokes of characters. Also includes puzzles that involve interpreting numeric digits as letters.
\end{itemize}

\subsection{Other Categories}
\begin{itemize}
    \item \textbf{Initials and Abbreviations:} Puzzles that involve initials, acronyms, or abbreviations. Includes puzzles in which the initials of two phrases must match; puzzles that involve acronyms or abbreviations of words; and puzzles that involve the symbols of elements on the periodic table.
    \item \textbf{Tricky:} Puzzles that involve a trick. Often, some phrase in the puzzle description must be interpreted in an unusually literal manner or in an unusual way.
    \item \textbf{Mathematical:} Puzzles that involve numbers, mathematical operations, or geometry. This includes puzzles that involve reasoning about combinatorics or probability, involve mathematical notation, require arithmetic, or involve geometric shapes (that are not character symbols).
    \item \textbf{Riddle:} Puzzles that are riddles. Defining a riddle is challenging. However, the language used to describe these puzzles is noticeably different from other puzzles. Usually, wordplay is required but not mentioned overtly.
    \item \textbf{Roman Numerals:} Puzzles that involve knowledge of roman numerals.
    \item \textbf{Synonyms and Antonyms:} Puzzles that state that portions of the answer are synonyms, near-synonyms, or antonyms of each other.
    \item \textbf{Other Wordplay:} Puzzles that involve an element of wordplay and do not fit into any non-knowledge category.
\end{itemize}

\subsection{Knowledge Categories}
\begin{itemize}
    \item \textbf{Geography:} Puzzles that require knowledge of geographic locations.
    \item \textbf{Sports and Games:} Puzzles that require knowledge of sports and games, sports teams, or athletes.
    \item \textbf{Pop Culture:} Puzzles that require knowledge of popular culture. Includes movies, television, celebrities, song titles, musicals, or entertainers.
    \item \textbf{Art:} Puzzles that require knowledge of artists, works of art, literature, or classical music.
    \item \textbf{Politics:} Puzzles that require knowledge of politicians or people of political significance.
    \item \textbf{History:} Puzzles that require knowledge of history.
    \item \textbf{General:} Puzzles that require general knowledge that does not belong to any above categories.
\end{itemize}

\section{Hardness of Challenges for Humans}
\label{human-hardness}

The \benchmark{} benchmark is based on the NPR Sunday Puzzle, which is a segment of the show \emph{Weekend Edition}. The show is broadcast on radio and streamed online and is estimated to reach over three million listeners every week~\citep{bryan:weekly-faq}. The show hosts do not consistently report how many solutions they receive. However,  when a challenge received approximately 1,500 answers, that was considered notably high~\citep{npr:high-count}.

In contrast, only four people solved the following challenge--perhaps an all-time low--and all four wrote computer programs to do so in 1995~\citep{renner:npr1995}. \emph{This challenge is not part of our benchmark.}

\challengeanswer{Take the two words, palmistry and behind. Together, they contain 15 letters without repetition. What are the two words in the Merriam Webster Tenth Collegiate Dictionary that have the most letters without any repetition?}{gunpowdery and blacksmith}

While the answer is correct, it is notable that the example in the question is wrong (the two words do repeat the letter `i`). The listeners were able to ignore this small error and understand the intent of the challenge.

\section{On-Air Challenges}
\label{on-air-challenges}

For the on-air challenges, each puzzle description applies to multiple challenge-answer pairs. The listeners are first presented with the description, followed by a list of questions where they are tasked with determining the answers on-air based on the provided description. The challenges are designed to be solvable in seconds and require less reasoning comparing to the off-air challenges.

We report the performance of GPT-4o and o3-mini on the on-air questions \citep{zhao:puzzleqa}. GPT-4o score 65.6\%, and o3-mini score 77.7\%, with a performance gap of 12\% - smaller than the 31\% gap observed for the off-air challenges. The significant differences in both the model performances and the relative improvement from model with enhanced reasoning ability suggest that our off-air puzzles require more reasoning.

\newcommand{\onairtwo}[5]{%
  \refstepcounter{challengecounter}%
  \begin{center}%
    \small%
    \renewcommand{\arraystretch}{1.5}%
    \noindent
    \makebox[\linewidth][l]{%
      % Inner table of fixed width
      \makebox[0pt][l]{%
        \begin{tabular}{|p{0.42\textwidth}|} 
          \hline
          \textbf{Description:} #1 \\ \hline
          \textbf{Question:} #2 \\ \hline
          \textbf{Answer:} #3 \\ \hline
          \textbf{Question:} #4 \\ \hline
          \textbf{Answer:} #5 \\
          \hline
        \end{tabular}%
      }%
      % Counter pushed to the far right
      \hfill
      (\thechallengecounter)%
    }%
  \end{center}%
}

This is one on-air challenge requiring knowledge on Rearrange Letters and Synonyms. Both o3-mini and GPT-4o answered correctly.

\onairtwo{Given a four-letter word and a six-letter word, rearrange the letters of one of them to get a synonym of the other}{peek, retain}{keep}{clam, serene}{calm}

Challenge~\ref{challenge:off-air1} is a similar off-air challenge that require Anagram and Synonym understanding. Compared to the on-air challenge, the off-air challenge does not provide clues regarding the synonyms, and the listeners are tasked to test through U.S. cities and rearrange their letters until a reasonable answer appears. Both o3-mini and GPT-4o failed on this challenge. Out of all models we benchmarked, only o1 and R1 succeeded.

\challengeanswer[challenge:off-air1]{Name a major U.S. city in 10 letters. If you have the right one, you can rearrange its letters to get two 5-letter words that are synonyms. What are they?}{Sacramento \texttt{-->} scent, aroma}

\newcommand{\onair}[4]{%
  \refstepcounter{challengecounter}%
  \begin{center}%
    \small%
    \renewcommand{\arraystretch}{1.5}%
    \noindent
    \makebox[\linewidth][l]{%
      \makebox[0pt][l]{%
        \begin{tabular}{|p{0.42\textwidth}|p{0.25\textwidth}|p{0.25\textwidth}|} 
          \hline
          \textbf{Description:} #1 \\ \hline
          \textbf{Question:} #2 \\ \hline
          \textbf{Answer:} #3 \\ 
          \hline
        \end{tabular}%
      }%
      \hfill
      (\thechallengecounter)%
    }%
  \end{center}%
}

Below is one on-air challenge with wordplay on initials/abbreviation. Both o3-mini and GPT-4o answered correctly.

\onair{Every answer today is the name of a famous person whose first initial and last name, in order, spell a word. For example, take Benjamin Rush, signer of the Declaration of Independence. The B of Benjamin + his last name spells BRUSH. I'll give you clues to the parts. You give me the names}{Singer with the group Hole — garlic bulb}{Courtney Love, clove}

Challenge~\ref{challenge:off-air2} is a similar off-air challenge that require Initials/abbreviation and Anagram understanding. However, it provides less clues about the person and about the field of renown, thus imposing a larger search space. o3-mini answered correctly, but GPT-4o failed.

\challengeanswer[challenge:off-air2]{Name a famous person with the initials M.C. The first initial and last name anagram to the person's field of renown. What is it?}{Michael Caine \texttt{-->} CINEMA}

% \begin{figure}
% \centering
% \includegraphics[width=0.45\textwidth]{figs/cdf_10.pdf}
% \caption{CDF of Success Rate Across 10 Completions vs. Cumulative Challenge Count}
% \label{fig:cdf_10}
% \end{figure}
% \section{Best-of-N Sampling}
% \label{best-of-n}

% We examine whether resampling improves model performance. We generate 10 completions per challenge using both R1 and Gemini-FT. For each challenge, we determine the percentage of correct responses from the 10 completions. The cumulative distribution function (CDF) of these percentages is shown in Figure~\ref{fig:cdf_10}. When success is defined as obtaining at least one correct response, R1 succeeds in 62\% of challenges and Gemini-FT in 43\%. Both models improve significantly over their average accuracy. Notably, R1 achieves performance comparable to that of O1.

\section{Are Certain Types of Challenges Harder than Others?}
\label{challenge-type-difficulty}

% \begin{figure}

% \centering
% \begin{subfigure}{0.45\textwidth}
% \includegraphics[width=\textwidth]{figs/knowledge_accuracy_distribution.pdf}
% \caption{Accuracy across knowledge categories}
% \label{fig:knowledge_category}
% \end{subfigure}
% \begin{subfigure}{0.45\textwidth}
% \includegraphics[width=\textwidth]{figs/wordplay_accuracy_distribution.pdf}
% \caption{Accuracy across puzzle types}
% \label{fig:puzzle_type}
% \end{subfigure}
% %

% \caption{Model accuracy across puzzle types and knowledge categories}
% \label{fig:type_category}
% \end{figure}

\begin{figure}
\centering
\includegraphics[width=0.5\textwidth]{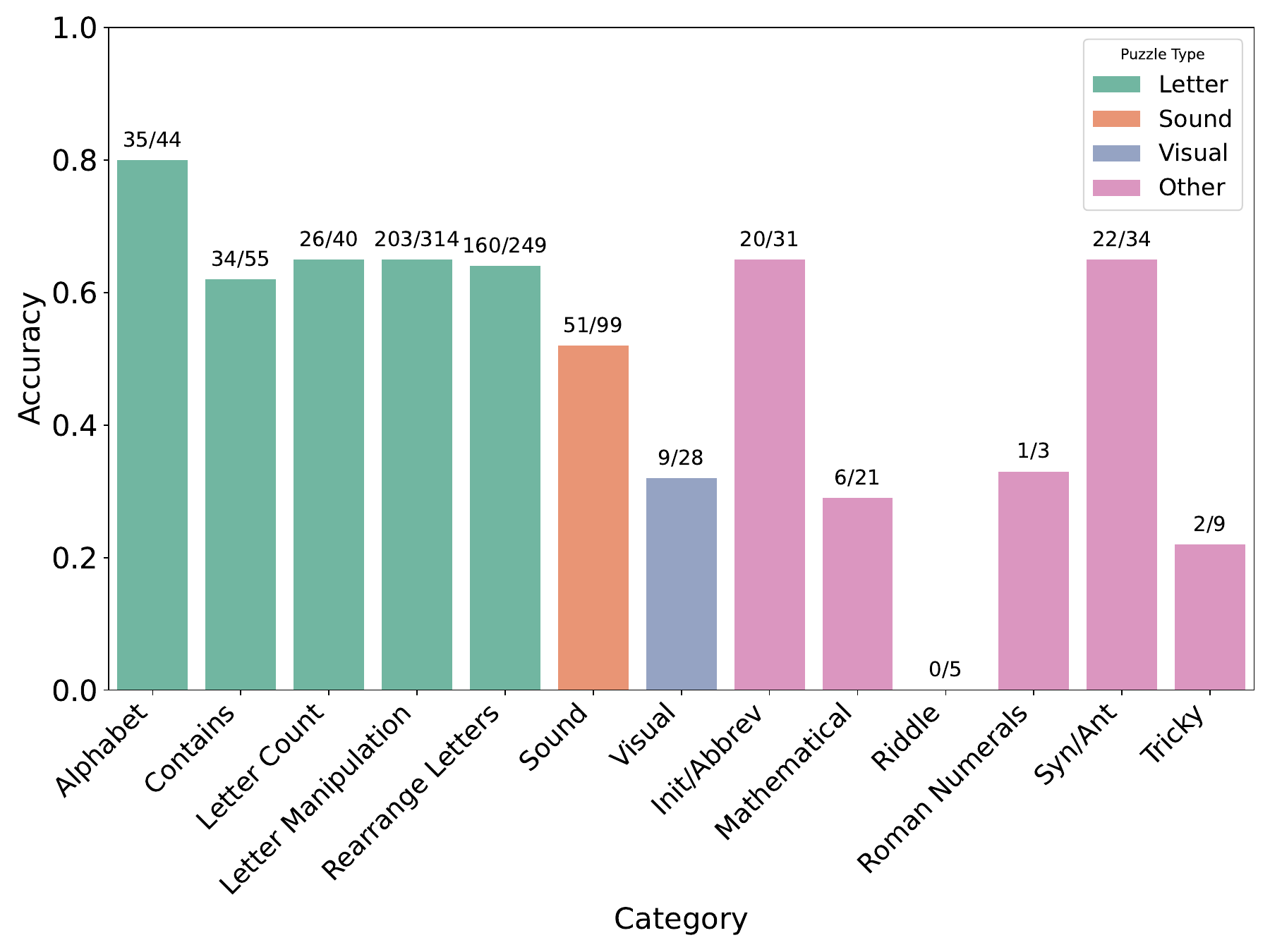}
\caption{o1's accuracy across puzzle categories}
\label{fig:o1_puzzle_category}
\end{figure}

We encoded the NPR challenges by the kind of wordplay and knowledge they involve, specified in Section \ref{coding-guideline}.
\cref{fig:type_category} shows the accuracy of the models across puzzle types and across knowledge categories. 
\cref{fig:knowledge_category} shows that most models achieve similar performance across knowledge categories. There are no specific knowledge categories that models particular excel or struggles in.

To test whether puzzle types significantly impact the performance of each model, we fit a multivariate multiple regression model. \footnote{We determine statistical significance using a p-value threshold of 0.05.} Both o1 and o3-mini-high perform better in the Letter category. In particular, puzzles that involve understanding the order of letters in the alphabet are the easiest for the models o1, o1-mini, o3-mini, o3-mini-high and Gemini-FT. \cref{fig:o1_puzzle_category} show o1's decline in performance when engaging with Mathematical, Riddle, Tricky puzzles, Visual puzzles including Crosswords and Character Properties, and Sound puzzles on rhyming and phonetic similarity. All other models observe no statistically significant variation in performance across puzzle types.

\section{Puzzle Solving Strategy as Reasoning Length Increases}
\label{candidates}

We investigate how reasoning models find potential answers or \emph{candidates} for a puzzle. We focus on the thinking output of reasoning models Gemini-FT and R1. We extract potential candidates from the model generation using two different methodologies. The first method is simple regex extraction, where we assume that a candidate is any phrase enclosed within a double quote (``). Note that this method is imprecise, since models do not necessarily provide candidates within double quotes. To improve upon the precision of the first method, we also use a second method involving a transformer-based structured retrieval model, NuExtract-1.5 \citep{cripwell2024nuextract}. We finetune the retrieval model on hand-annotated training pairs of generations and candidate labels, taken from the output of another model not in our test set. This step specializes NuExtract-1.5 on our specific task and prompt format. Finally, we use our finetuned retrieval model to extract candidates from R1 and Gemini-FT generations. 

In \cref{fig:candidate_search}, we plot the results of our candidate extraction. We find that beyond a certain length $l$, models cease to find new candidates for a puzzle. While the value of $l$ varies between our candidate extraction methods, in both cases, the $y$-axis eventually converges into an asymptote at 100\%. The regex-method can be thought of as providing a looser lower bound to the number of candidates. This means that even when given more thinking time, these models will not produce any new candidates. This confirms our previous finding that model accuracy reaches a plateau after n-reasoning tokens. Intuitively, the reason why this happens is because models cease to find new reason
ing branches that discover novel candidates.

\section{Dataset and Code Availability}
\label{artifact-url-and-license}
The code and dataset for this submission is publicly available and licensed under an MIT license. The dataset is derived from NPR Sunday Puzzle challenges and is subject to NPR’s copyright and licensing terms.

\ifanon
The URL is omitted for review. 
\else
The code and data for this paper are available at \url{https://huggingface.co/datasets/nuprl/reasoning-weekly}.
\fi

\section{Use of AI Assistants}\label{use-of-ai-assistants}

Some code for this paper was written with AI assistants enabled.

\section{Computing Resources}
\label{computing-resources}
The computational experiments for this paper were conducted with less than 100 hours of A100 GPU time, along with approximately 30 million tokens processed via hosted models.

\clearpage

\section{Checklist}

A2 Potential Risks: 
Did you discuss any potential risks of your work? [Yes/No/NA]
\textbf{NA}

A2 Elaboration:
For yes, provide a section number. For no, justify why not.

B Use Or Create Scientific Artifacts:
Did you use or create scientific artifacts? [Yes/No]
\textbf{Yes}

B1 Cite Creators Of Artifacts:
Did you cite the creators of artifacts you used? [Yes/No/NA]
\textbf{Yes}

B1 Elaboration:
For yes, provide a section number. For no, justify why not.
\textbf{Section \ref{sec:dataset}.}

B2 Discuss The License For Artifacts:
Did you discuss the license or terms for use and/or distribution of any artifacts? [Yes/No/NA]
\textbf{Yes}

B2 Elaboration:
For yes, provide a section number. For no, justify why not. \textbf{Section \ref{artifact-url-and-license}}

B3 Artifact Use Consistent With Intended Use:
Did you discuss if your use of existing artifact(s) was consistent with their intended use, provided that it was specified? For the artifacts you create, do you specify intended use and whether that is compatible with the original access conditions? [Yes/No/NA]
\textbf{Yes}

B3 Elaboration:
For yes, provide a section number. For no, justify why not.
\textbf{Ethics Statement}

B4 Data Contains Personally Identifying Info Or Offensive Content:
Did you discuss the steps taken to check whether the data that was collected/used contains any information that names or uniquely identifies individual people or offensive content, and the steps taken to protect/anonymize it? [Yes/No/NA]
\textbf{NA}

B4 Elaboration:
For yes, provide a section number. For no, justify why not.

B5 Documentation Of Artifacts:
Did you provide documentation of the artifacts, e.g., coverage of domains, languages, and linguistic phenomena, demographic groups represented, etc.? [Yes/No/NA]
\textbf{Yes}

B5 Elaboration:
For yes, provide a section number. For no, justify why not.
\textbf{Section \ref{challenge-type-difficulty}.}

B6 Statistics For Data:
Did you report relevant statistics like the number of examples, details of train/test/dev splits, etc. for the data that you used/created? [Yes/No/NA]
\textbf{Yes}

B6 Elaboration:
For yes, provide a section number. For no, justify why not.
\textbf{Section \ref{sec:dataset}}

C Computational Experiments:
Did you run computational experiments? [Yes/No/NA]
\textbf{Yes}

C1 Model Size And Budget:
Did you report the number of parameters in the models used, the total computational budget (e.g., GPU hours), and computing infrastructure used? [Yes/No/NA]
\textbf{Yes}

C1 Elaboration:
For yes, provide a section number. For no, justify why not.
\textbf{Section \ref{computing-resources}}

C2 Experimental Setup And Hyperparameters:
Did you discuss the experimental setup, including hyperparameter search and best-found hyperparameter values?
[Yes/No/NA]
\textbf{Yes}

C2 Elaboration:
For yes, provide a section number. For no, justify why not.
\textbf{Section \ref{model-config}.}

C3 Descriptive Statistics:
Did you report descriptive statistics about your results (e.g., error bars around results, summary statistics from sets of experiments), and is it transparent whether you are reporting the max, mean, etc. or just a single run? [Yes/No/NA]
\textbf{Yes}

C3 Elaboration:
For yes, provide a section number. For no, justify why not.
\textbf{Sections \ref{sec:results} and \ref{challenge-type-difficulty}.}

C4 Parameters For Packages:
If you used existing packages (e.g., for preprocessing, for normalization, or for evaluation, such as NLTK, SpaCy, ROUGE, etc.), did you report the implementation, model, and parameter settings used? [Yes/No/NA]
\textbf{Yes}

C4 Elaboration:
For yes, provide a section number. For no, justify why not. \textbf{Section \ref{model-config}.}

D Human Subjects Including Annotators:
Did you use human annotators (e.g., crowdworkers) or research with human subjects? [Yes/No/NA]
\textbf{No}

D1 Instructions Given To Participants:
Did you report the full text of instructions given to participants, including e.g., screenshots, disclaimers of any risks to participants or annotators, etc.? [Yes/No/NA]
\textbf{NA}

D1 Instructions Given To Participants:
Did you report the full text of instructions given to participants, including e.g., screenshots, disclaimers of any risks to participants or annotators, etc.? [Yes/No/NA]
\textbf{NA}

D1 Elaboration:
For yes, provide a section number. For no, justify why not.

D2 Recruitment And Payment:
Did you report information about how you recruited (e.g., crowdsourcing platform, students) and paid participants, and discuss if such payment is adequate given the participants' demographic (e.g., country of residence)? [Yes/No/NA]
\textbf{NA}

D2 Elaboration:
For yes, provide a section number. For no, justify why not.

D3 Data Consent:
Did you discuss whether and how consent was obtained from people whose data you're using/curating (e.g., did your instructions explain how the data would be used)? [Yes/No/NA]
\textbf{NA}

D3 Elaboration:
For yes, provide a section number. For no, justify why not.

D4 Ethics Review Board Approval:
Was the data collection protocol approved (or determined exempt) by an ethics review board? [Yes/No/NA]
\textbf{NA}

D4 Elaboration:
For yes, provide a section number. For no, justify why not.

D5 Characteristics Of Annotators:
Did you report the basic demographic and geographic characteristics of the annotator population that is the source of the data? [Yes/No/NA]
\textbf{NA}

D5 Elaboration:
For yes, provide a section number. For no, justify why not.

E Ai Assistants In Research Or Writing:
Did you use AI assistants (e.g., ChatGPT, Copilot) in your research, coding, or writing? [Yes/No]
\textbf{Yes}

E1 Information About Use Of AI Assistants:
Did you include information about your use of AI assistants? [Yes/No/NA]
\textbf{Yes}

E1 Elaboration:
For yes, provide a section number. For no, justify why not.
\textbf{Section \ref{use-of-ai-assistants}.}
\end{document}